\documentclass[conference]{IEEEtran}
\IEEEoverridecommandlockouts
\usepackage{cite}
\usepackage{hyperref}
\usepackage{amsmath, amssymb, amsfonts}
\usepackage{algorithmic}
\usepackage{graphicx}
\usepackage{textcomp}
\usepackage[dvipsnames, svgnames, x11names]{xcolor}
\usepackage{amssymb}
\usepackage{booktabs}
\usepackage{multirow}
\usepackage{diagbox}

\usepackage[]{makecell}
\def\BibTeX{{\rm B\kern-.05em{\sc i\kern-.025em b}\kern-.08em T\kern-.1667em\lower.7ex\hbox{E}\kern-.125emX}}
\newcommand{\ours}{\texttt{FAIR-ESI}}
\renewcommand{\emph}[1]{\noindent
\textbf{#1.}}
\definecolor{Garnet}{RGB}{115, 0, 10} 
\definecolor{CitationTeal}{RGB}{0,128,128} 
\hypersetup{
    colorlinks=true, 
    linkcolor=blue, 
    citecolor=red, 
    urlcolor=green, 
}

\begin{document}
    \title{\ours: Feature Adaptive Importance Refinement for Electrophysiological Source Imaging {

    }
    \thanks{
    }
    \vspace{-1.0em}}
\author{\IEEEauthorblockN{Linyong Zou$^{\dag,1,2}$, Liang Zhang$^{\dag,1,2,3}$, Xiongfei Wang$^{\dag,4}$, Jia-Hong Gao$^{5}$, Yi Sun$^{2}$, Shurong Sheng$^{2}$, 
 Kuntao Xiao$^{2}$, \\ Wanli Yang$^{2}$, Pengfei Teng$^{4}$, Guoming Luan$^{*,4}$, Zhao Lv$^{*,6}$ and Zikang Xu$^{*,2}$ \\}
    $^{1}$ School of Artificial Intelligence, Anhui University, Hefei 230601, China.\\
    $^{2}$ Anhui Province Key Laboratory of Biomedical Imaging and Intelligent Processing, \\Institute of Artificial Intelligence, Hefei Comprehensive National Science Center, Hefei 230088, China.\\
    $^{3}$ School of Automation, Northwestern Polytechnical University, Xian 710072, China. \\ 
    $^{4}$ Department of Neurosurgery, Beijing Key Laboratory of Epilepsy, Sanbo Brain  Hospital Capital Medical University,\\ Beijing, China.\\
    $^{5}$ McGovern Institute for Brain Research, Peking University, Beijing, China.\\
    $^{6}$ Anhui Province Key Laboratory of Multimodal Cognitive Computation, School of Computer Science and Technology, \\Anhui University, Hefei 230601, China.\\ 
\IEEEauthorblockA{$^{\dag}$ Equal contribution, and $^{*}$ Corresponding authors.} 
\IEEEauthorblockA{luangm@ccmu.edu.cn, kjlz@ahu.edu.cn, zikangxu@mail.ustc.edu.cn} 
}

    \maketitle

    \begin{abstract}
        An essential technique for diagnosing brain disorders is electrophysiological source imaging (ESI). While model-based optimization and deep learning methods have achieved promising results in this field, the accurate selection and refinement of features remains a central challenge for precise ESI. This paper proposes \ours, a novel framework that adaptively refines feature importance across different views, including FFT-based \textit{spectral} feature refinement, weighted \textit{temporal} feature refinement, and self-attention-based \textit{patch-wise} feature refinement. Extensive experiments on two simulation datasets with diverse configurations and two real-world clinical datasets validate our framework's efficacy, highlighting its potential to advance brain disorder diagnosis and offer new insights into brain function.
    \end{abstract}


    \section{Introduction}

    Brain disorders, including epilepsy, Parkinson's disease,
    and depressive disorder, have brought substantial suffering for a large number of patients all over the world every year~\cite{world2006neurological}. Precise localization of the deep-seated pathological sources of these psychiatric disorders in the brain is critical for developing targeted neuromodulation therapies~\cite{laumann2023precision}. However, traditional localization methods, such as stereoelectroencephalography (sEEG), require intracranial electrode implantation, which is expensive and high-risk~\cite{brunoni2018noninvasive}. Due to the development of neuroimaging technologies, some researchers acquire non-invasive electrophysiological signals from patients' scalps, such as magnetoencephalography (MEG) and electroencephalography (EEG), followed by source localization algorithms to estimate intracranial activity~\cite{ref_inverse_problem}. This strategy substantially reduces both surgical costs and patients' discomfort.

\begin{figure}[htbp]
        \vspace{-1.0em}
        
        \centering
        \includegraphics[width=\linewidth]{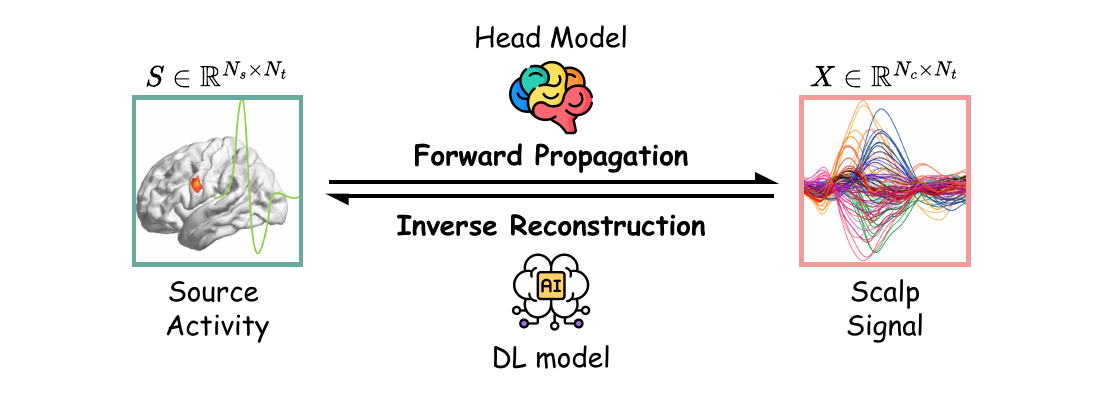}
        \caption{The forward propagation and inverse reconstruction problem of ESI.}
        \label{fig:background}
        \vspace{-0.5em}
    \end{figure}

    Electrophysiological source imaging (ESI)~\cite{ref_inverse_problem} refers to the process of reconstructing brain source activities from scalp signals, as illustrated in Fig.~\ref{fig:background}. This technique is inherently an \textit{inverse problem} that reverses the forward propagation model of electromagnetic fields from neural sources to scalp sensors~\cite{ref_inverse_problem}. However, due to the sparse spatial sampling of electromagnetic sensors on the scalp, ESI becomes an ill-posed problem with non-unique solutions~\cite{ref_XDLESI}.

    Over the past decades, researchers have attempted to address this ill-posed problem through optimization-based or data-driven algorithms. Traditional optimization-based approaches solve the ESI inverse problem by imposing regularization constraints on the source space to ensure solution uniqueness. While priors in these methods enhance localization accuracy~\cite{ref14,ref15,ref17,ref18}, formulating appropriate regularization constraints remains challenging~\cite{ref34}. Moreover, these methods suffer from two inherent limitations: (1) most of the methods assume static sources, failing to capture transient neural dynamics; (2) the head geometry and tissue conductivity variations require case-specific forward modeling.

    To tackle the shortcomings of the traditional methods, some researchers try to solve the ESI problem by adopting deep learning (DL)-based algorithms. By leveraging massive simulated data to tune neural network parameters, DL-based methods can learn more flexible and realistic reconstruction functions via learning comprehensive spatiotemporal features directly from source-scalp signal pairs~\cite{ref_DeepSIF}.

    However, current DL-based algorithms tend to treat the following characteristics wrongly when extracting features from the input signals, which leads to inaccurate localization of the source signal: (1) {\textit{frequency components}}: Brain source activities exhibit distinct electrophysiological signatures across frequency bands, providing discriminative features for ESI~\cite{ref_PNAS_problem}. However, most DL models disregard these spectral characteristics, leading to suboptimal feature representation; (2) {\textit{temporal samples}}: Source spikes induce transient electrophysiological responses on the scalp, where specific time points strongly correlate with underlying neural activity while others contain volume conduction artifacts. Nevertheless, many DL algorithms process entire signal fragments indiscriminately, reducing feature SNR; and (3) {\textit{Sub-patches}}: Processing full scalp signals introduces spatiotemporal contamination-sub-regions containing activity from preceding/succeeding neural events are erroneously used for source localization. This results in misattribution of signal origins and degraded spatial accuracy.

    To solve these issues, in this paper, we propose \ours, which improves the precision of ESI by adaptively refining the importance of the extracted features from three different views:

    \begin{enumerate}
        \item From the view of \textit{spectral}: \ours~adopts a Fast Fourier Transform (FFT) \cite{ref_fft} based feature refinement strategy which adaptively selects important features in the frequency domain, and then converts the refined feature into the temporal domain;

        \item From the view of \textit{temporal}: \ours~applies weighted summation to fuse features in the temporal domain, which can reduce noise while preserving details of the features;

        \item From the view of \textit{patch-wise}: \ours~selects the key patch based on energy and enhances it via self-attention, and then this salient feature is propagated across all patches to refine patch-wise features.
    \end{enumerate}

    Furthermore, paired source-scalp signals are scarce owing to the challenges in simultaneous signal acquisition, which bring difficulties for model training. To solve this problem, we leverage neural mass models (NMMs) \cite{ref_nmms}, which are used to generate brain source activities by numerical simulation, to generate abundant paired data with diverse configurations. As described above, the contributions of this work are as follows:

    \begin{enumerate}
        \item An NMM-based method is adopted to simulate paired source-scalp signals with diverse Signal-to-Noise Rate (SNR), number of sources, and number of neighborhood extents, which are used to train the DL-based ESI model;

        \item A DL model, named \ours, which adaptively refines feature importance from three views including \textit{spectral}, \textit{temporal}, and \textit{patch-wise}, is proposed for precise ESI from scalp signals;

        \item Extensive experiments on two simulation datasets and two real-world clinical datasets prove the effectiveness and generalizability of the proposed model, which also provides potential for automatic brain disorder diagnosis, especially for epilepsy.
    \end{enumerate}

    \color{black}

    \section{Methodology}

    \subsection{Preliminaries}
    Let $S \in \mathbb{R}^{N_s \times N_t}$ denote the brain source activity in $N_{s}$ regions of interest (ROIs) at $N_{t}$ time points, and $X \in \mathbb{R}^{N_c \times N_t}$ denote the corresponding scalp fragments (EEG/MEG) in $N_{c}$ channels at $N_{t}$ time points. The forward problem of ESI can be expressed as $X = G\cdot S + N, \label{fwd_equation}$
    where $G \in \mathbb{R}^{N_c \times N_s}$ is the lead-field matrix derived from brain structure, which measures the coupling between a unit current dipole at a specific cortical location and the resultant electromagnetic field measured at each MEG/EEG sensor, and $N \in \mathbb{R}^{N_c \times N_t}$ denotes the measurement noise.

    The inverse problem of ESI, which aims to reconstruct the brain source activity $S$ from the observation of EEG/MEG signals, is an ill-posed problem due to the underdetermined nature of the mapping from source space to sensor space. To address this, we aim to learn an end-to-end mapping function $P$ that reconstructs the brain source activity $S$ from the EEG/MEG signals $X$ via DL, i.e., $X \xrightarrow{P}S$.
    \subsection{Paired Data Simulation \& Patch Data Generation}

    Because the paired scalp signals $X$ and brain source activities $S$ are precious and rare, following~\cite{ref_DeepSIF}, we employ a neural mass model (NMM) to simulate the paired data, which is then used to train our DL model. By adjusting the parameters of the NMM-based Jansen-Rit algorithm~\cite{ref_nmms}, we generate a wide range of simulated data with different signal-to-noise rates (SNRs, $r$), number of sources ($N_s$), and spatial extents ($N_n$). Besides, by adopting different head models (Fsaverage2020~\cite{fischl1999high} and ICBM152~\cite{mazziotta1995probabilistic}) and lead-field metrics, we can generate both MEG and EEG signals from source activities. This approach allows us to create a diverse dataset that captures the dynamics of neural populations and provides a realistic representation of brain activity.


    Unlike previous methods that directly process the entire scalp signal to extract features representing the whole fragment, or splitting the signal into channel-wise segments and generate features for each channel separately, we propose to generate overlapping patches for a more comprehensive patch-wise representation of the signal. This approach allows us to capture patch-wise patterns and enhance the model's ability to learn from the data.

    
    For each scalp fragment $X \in \mathbb{R}^{N_c \times N_t}$, we partition it into channel-independent temporal-overlapped segments, resulting in $P\in \mathbb{R}^{N_c \times N_p}$ patches, where $N_{p}$ denotes the number of segments for each channel.
    \subsection{\ours: Feature Adaptive Importance Refinement}

    To address the aforementioned challenges in the ESI inverse problem, we propose a novel DL algorithm, i.e. \ours, which is achieved by adaptively refining the features from the scalp fragments considering the feature importance from \textit{spectral}, \textit{temporal}, and \textit{patch-wise} perspectives.

    \subsubsection{FFT-based Spectral Feature Refinement}

    From a \textit{spectral} point of view, we follow the findings in~\cite{ref_PNAS_problem}, which hint that some of frequency components are important while usually ignored in the ESI inverse problem. Besides, the background activity in the frequency domain is a common distraction that prevents the model from learning the important features.

    For each patch $P(i, j)$, \ours~first applies the Fast Fourier Transform (FFT) to bring the temporal signal into the frequency domain. Then, the real and imaginary parts of the FFT results are processed separately using a temperature-scaled softmax function to refine the features and remove irrelevant noise, i.e., $\mathcal{T}(x , \tau) = \frac{\exp(x / \tau)}{\sum \exp(x / \tau)}$. After that, an Inverse FFT (IFFT) is adopted to reconstruct the temporal signal from the two components, as shown in Equ.~\eqref{eq:spectral_feature}.
    \begin{equation}
        P^{*}_{S}= \text{IFFT}(P^{*}_{\text{RE}}, P^{*}_{\text{IM}}) = \text{IFFT}(\mathcal{T}(P_{\text{RE}}, \tau), \mathcal{T}(P_{\text{IM}}, \tau)),\label{eq:spectral_feature}
    \end{equation}
    where $[P_{\text{RE}}, P_{\text{IM}}]=\text{FFT}(P)$ denote the real and imaginary parts of the FFT results, respectively.

    \subsubsection{Weighted Temporal Feature Refinement}
    However, as shown in Fig.~\ref{fig:model_overview}~(b), the refinement in the spectral domain may result in over-smoothing of the signal, which can cause loss of details in the temporal domain. To address this, we further refine the features from the view of the \textit{temporal} domain.

    Specifically, we first use a temperature-scaled softmax function to refine the input patch, i.e., $P^{*}_{T}= \mathcal{T}(P, \tau)$. Then, a weighted-summation operation is applied to fuse the refined spectral features $P^{*}_{S}$ and the refined temporal features $P^{*}_{T}$, which is expressed as $P^{L}= \alpha \cdot P^{*}_{S}+ (1 - \alpha) \cdot P^{*}_{T}$, where $\alpha$ is a hyper-parameter that controls the balance between the two components.

   

    \subsubsection{Self-Attention-based Patch-wise Feature Refinement}

    By the above two steps, we have refined the features for each patch. However, in the whole scalp fragment $P$ with $N_{c}\times N_{p}$ patches, not all patches are of equal importance. For example, some patches may contain activated spikes which are critical for source localization, while others may represent resting states which are less relevant. To address this, we propose a cross-attention-based global feature refinement strategy to adaptively fuse the features from all patches.

    As shown in Fig.~\ref{fig:model_overview}~(c), for the patches $P^{L}(i,j), j \in[1,N_{p}]$ of the $i$-th channel, we first compute the element-wise square of each patch. Then, the key patch of the $i$-th channel, $P^{L}(i, p)$, is identified as the one with the maximum energy, as shown in $E(i,j) = P^{L}(i,j) \odot P^{L}(i,j), \forall j\in[1, N_{p}]$,
    where $\odot$ denotes the hadamard product, and $p = \underset{j}{\arg\max}~ E(i,j)$ is the index of the key patch. Then, we apply self-attention~\cite{ref_self_attention} to $p^{L}(i,j)$ and propagate the attention weights across $P^{L}$:
    \begin{align}
        Q = W_{Q}\cdot P_{i,p}^{L}, K & = W_{K}\cdot P_{i,p}^{L}, V = W_{V}\cdot P_{i,p}^{L},     \\
        \text{attention}              & = \text{Softmax}(\frac{Q\cdot K^{T}}{ \sqrt{d}}) \cdot V, \\
        P^{A}                         & = P^{F}\oplus \text{attention},
    \end{align}
    where $d$ is the dimension of $Q$ and $K$, $W_{Q},W_{K},W_{V}$ are the projection metrics, and $\oplus$ denotes concatenate operation. Finally, the attention-enhanced features $P^{A}$ are sent to two sequential convolutional blocks (Conv2D-ELU-LayerNorm, and TransposeConv2D-ELU-LayerNorm) to obtain the refined features $P^{O*} = \text{TransposeConvBlk}(\text{ConvBlk}(P^{A}))$.

    After repeating this feature adaptive importance refinement procedure for $N$ times, the final output feature $P^{O}$ is used to reconstruct the source activity.

    \subsubsection{Source Activity Reconstruction}

\begin{figure*}[htbp]
    \vspace{-2.0em}
        \centering
        \includegraphics[width=0.8\linewidth]{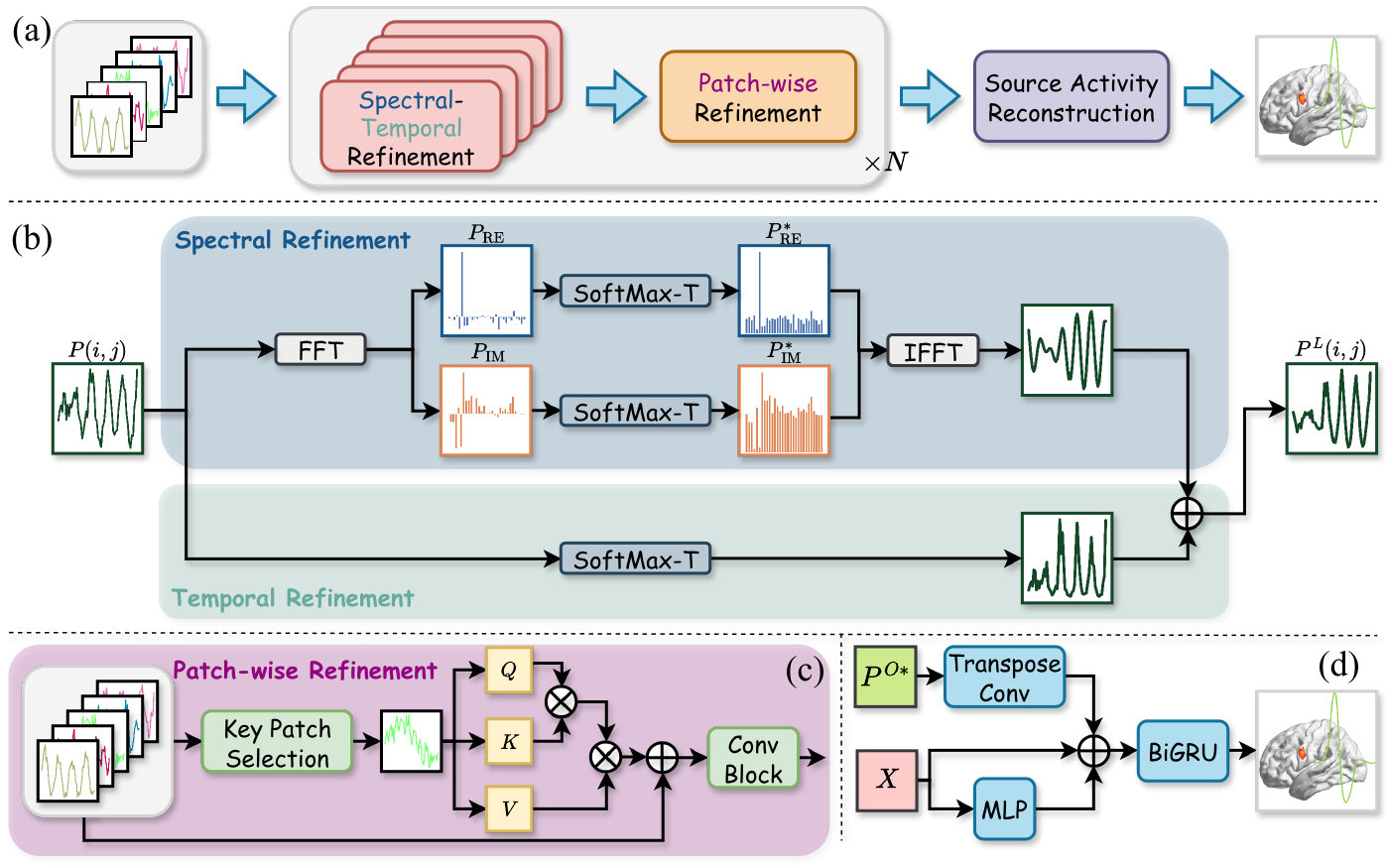}
        \caption{Overview of \ours. (a) The pipeline of the proposed method, which contains three views of feature refinement. (b) Details about spectral-temporal feature refinement. Note that $P^{*}_{\text{RE}}$ and $P^{*}_{\text{IM}}$ are in log-scale for better visualization. (c) Details about patch-wise feature refinement. (d) Details about source activity reconstruction. $P^{O*}$ and $X$ are the output of the feature refinement module and the input scalp signal, respectively. }
        \label{fig:model_overview}
        \vspace{-1.0em}
    \end{figure*}

    Following~\cite{ref_DeepSIF,ref_SSINet}, the refined features, $P^{O*}$, along with the input scalp signal, $X$, are fed into the source activity reconstruction module to predict the source activity $\hat{S}$, as shown in Fig.~\ref{fig:model_overview}~(d).

    For $P^{O*}$, we employ the same transpose convolutional block to upsample its spatial dimension to match that of the source space, while simultaneously restoring the temporal dimension to its original length.

    To preserve essential features in deeper layers, the original input $X$ is projected to the source space dimension via a multi-layer perceptron (MLP). This projection is augmented with a residual branch to mitigate information loss.

    Finally, $\hat{S}$ is generated by summing the three components, followed by processing through a bidirectional gated recurrent unit (BiGRU) network~\cite{li2019bigru}, as illustrated by Equ.~\eqref{eq:source_reconstruction}.
    \begin{equation}
        \hat{S}= \text{BiGRU}(\text{TransposeConv}(P^{O*})+ \text{MLP}(X) + X).\label{eq:source_reconstruction}
    \end{equation}
    \subsubsection{Loss Function}

    The model parameters are optimized iteratively using mean squared error (MSE) loss $\mathcal{L}$ between the predicted source activity $\hat{S}$ and the ground truth source map $S$, as shown in $  \mathcal{L}= \frac{1}{N_{s}}\| \hat{S}- S \|_{F}^{2}\label{eq:loss_function}$.
    




    \section{Experiments and Results}

    \subsection{Datasets}
    Both simulation datasets, named as SimMEG and SimEEG, and real-world datasets, including CMR \cite{ref_cmr} and Localize-MI~\cite{ref_HDEEG}, are used to evaluate the performance of the proposed model.

    \subsubsection{Simulation Datasets}
    Following~\cite{ref_DeepSIF}, the Jansen-Rit model is adopted to generate diverse paired data. The details about the two simulation datasets are shown in Table~\ref{tab:simulate_dist}. For each configuration, we generate 47,904 samples and split them into train, validation, and test sets with a ratio of 10:1:1.

    \begin{table}[htbp]
    \vspace{-1.0em}
        \centering
        \caption{Configurations of the simulation datasets.}
        \label{tab:simulate_dist} \resizebox{\linewidth}{!}{
        \begin{tabular}{lrr}
            \toprule \textbf{Configurations} & \textbf{SimMEG}                 & \textbf{SimEEG}                           \\
            \midrule Head Model              & Fsaverage2020~\cite{fischl1999high} & ICBM152~\cite{mazziotta1995probabilistic} \\
            $\#$ of Brain regions            & $998$                           & $998$                                     \\
            $\#$ of Time points              & $500$                           & $260$                                     \\
            $\#$ of Scalp channels           & $306$                           & $256$                                     \\
            $\#$ of Sources                  & {$1, 2, 3$}                     & {$1$}                                     \\
            $\#$ of Neighborhood Extents     & {$2, 3$}                     & {$2$}                                     \\
            SNR (dB)                         & {$-5, 5, 15$}                   & {$5$}                                     \\ 
            \bottomrule
        \end{tabular}
        }
    \vspace{-1.0em}
    \end{table}

    \subsubsection{Real-world Clinical Datasets}
    One private MEG dataset and one publicly-available EEG dataset are used in this study.

    \emph{CMR Dataset} The private CMR dataset \cite{ref_cmr} contains 306-channel MEG recordings and corresponding MRIs from 13 epilepsy patients. Our study on the CMR dataset was approved by the relevant institutional review boards.
    Each patient's MEG recordings are complemented by simultaneous 216-channel stereoelectroencephalography (sEEG) data, which provides definitive ground-truth localization for both seizure-onset zones and interictal epileptiform discharge zones across all brain regions. These labels specifically identify electrophysiologically active areas adjacent to sEEG electrode contacts.

    \emph{Localize-MI Dataset} The Localize-MI dataset \cite{ref_HDEEG} is an open resource featuring high-density EEG data from 61 sessions involving 7 subjects with drug-resistant epilepsy, designed for assessing source localization methods. This dataset provides 256-channel EEG signals at a sampling frequency of 8 kHz, along with spatial coordinates for stimulating brain contacts, which serve as the ground truth for source localization.

    \subsection{Experiment Configurations}

    The proposed \ours~is optimized using Adam, with a weight decay of 1e-5. The initial learning rate is set as 1e-4 and is dynamically reduced by half whenever the test error plateaus until reaching its minimum. The patch length $l$ is set as 16, with an overlap of 8. The temperature of softmax is set as 0.1. The number of \texttt{FAIR} blocks is set as $N=1$. All experiments are conducted on a Linux server with 8x NVIDIA RTX 4080 Graphics and repeated three times to avoid random errors.

    To evaluate the performance of \ours, we also implement eight state-of-the-art ESI algorithms in three categories: (1) \textbf{Model-based methods:} sLORETA~\cite{ref15}, Champagne~\cite{ref_champagne}; (2) \textbf{Data-driven methods:} ConvDip~\cite{ref_ConvDip}, DeepSIF~\cite{ref_DeepSIF}, SSINet~\cite{ref_SSINet}, ADMM-ESI~\cite{ref_ADMMESI}; (3) \textbf{General time-series methods:} Catch~\cite{ref_Catch}, FreEFormer~\cite{ref_FreEformer}. Following previous studies, five metrics are adopted to inspect the performance of the models, including precision~\cite{ref_DeepSIF}, recall~\cite{ref_DeepSIF}, localization error (LE)~\cite{ref_DeepSIF}, spatial dispersion (SD)~\cite{ref_DeepSIF}, and normalized mean squared error (nMSE) ~\cite{ref_SSINet}. Specifically, LE measures the minimum distance between the estimated source location and the ground truth region, and SD is used to describe the spatial blurring relative to the ground truth.

    \subsection{Quantitative Results}

    \emph{Results on SimMEG dataset} As shown in Table~\ref{tab:simmeg}, for most of simulation configurations, including different SNR, number of sources, and number of neighborhood extents, \ours~outperforms other state-of-the-art algorithms. This is followed by SSINet~\cite{ref_SSINet}, which extracts spatial and temporal features simultaneously from scalp fragments. Besides, for the configuration in which the noise is extremely high (SNR=-5 dB), all the other algorithms cannot achieve a precision higher than 80\%, while \ours~results in a precision of 83.99\%. This is due to the fact that \ours~conducts more meticulous feature refinements from diverse views. Besides, although SSINet obtains the highest recall in most of configurations, this occurs because SSINet prioritizes more active sources and facilitates the detection of a larger number of abnormal spikes, which leads to a lower precision.

    \begin{table*}
        [htbp]
        \vspace{-1.0em}
        \centering
        \caption{Results on SimMEG and SimEEG datasets. All experiments were repeated 3 times, mean $\pm$ std are reported. The \textbf{best} and \underline{second best} results are highlighted. $r , N_{s}, N_{n}$ denote the SNR, Number of sources, and number of neighborhood extents, respectively.}
        \label{tab:simmeg}
        \resizebox{\linewidth}{!}{
        \begin{tabular}{ccccccccccccc}
            \toprule \textbf{$r$}                            & \textbf{$N_{s}$}            & \textbf{$N_{n}$}            & \textbf{Metrics}        & \textbf{sLORETA}~\cite{ref15} & \textbf{Champange}~\cite{ref_champagne} & \textbf{ConvDip}~\cite{ref_ConvDip} & \textbf{DeepSIF}~\cite{ref_DeepSIF} & \textbf{SSINet}~\cite{ref_SSINet} & \textbf{ADMM-ESI}~\cite{ref_ADMMESI} & \textbf{Catch}~\cite{ref_Catch} & \textbf{FreEFormer}~\cite{ref_FreEformer} & \textbf{\ours}              \\ \bottomrule
           
            \multicolumn{13}{c}{ SimMEG dataset}  \\ 
            \midrule \multirow{5}{*}{\textbf{-5 }}           & \multirow{5}{*}{\textbf{1}} & \multirow{5}{*}{\textbf{2}} & Precision (\%)          & 8.15$\pm$9.39                     & 9.54$\pm$11.10                          & 31.09$\pm$40.61                     & \underline{79.25$\pm$19.19}         & 72.05$\pm$18.12                   & 61.68$\pm$17.12                      & 68.81$\pm$22.37                 & 60.87$\pm$22.62                           & \textbf{83.99$\pm$13.18}    \\
                                                             &                             &                             & Recall (\%)             & 18.79$\pm$18.45                   & 19.66$\pm$24.92                         & 20.56$\pm$30.39                     & 83.85$\pm$20.97                     & \textbf{92.87$\pm$12.27}          & 60.87$\pm$30.95                      & 78.99$\pm$24.51                 & 89.19$\pm$16.75                           & \underline{90.09$\pm$12.93} \\
                                                             &                             &                             & LE (mm)                 & 80.09$\pm$39.21                   & 66.18$\pm$36.98                         & \underline{2.37$\pm$4.22}           & 2.88$\pm$7.20                       & 3.03$\pm$3.97                     & 8.61$\pm$9.19                        & 3.50$\pm$4.30                   & 5.89$\pm$8.07                             & \textbf{1.96$\pm$6.04}      \\
                                                             &                             &                             & SD (mm)                 & 84.31$\pm$38.12                   & 76.12$\pm$41.83                         & 3.42$\pm$4.54                       & 4.18$\pm$7.70                       & \underline{3.35$\pm$4.13}         & 9.15$\pm$10.57                       & 4.14$\pm$4.54                   & 6.18$\pm$7.56                             & \textbf{2.55$\pm$6.45}      \\
                                                             &                             &                             & nMSE ($\times 10^{-4}$) & 13.32$\pm$4.59                    & 17.54$\pm$8.19                          & 6.30$\pm$4.20                       & 2.49$\pm$2.20                       & \underline{2.36$\pm$3.56}         & 6.17$\pm$8.40                        & 3.48$\pm$2.61                   & 4.48$\pm$6.34                             & \textbf{1.94$\pm$1.79}      \\
            \midrule \multirow{5}{*}{\textbf{5 }}            & \multirow{5}{*}{\textbf{1}} & \multirow{5}{*}{\textbf{2}} & Precision (\%)          & 13.19$\pm$16.02                   & 15.63$\pm$18.93                         & 50.11$\pm$33.25                     & \underline{82.20$\pm$16.40}         & 75.14$\pm$16.77                   & 68.59$\pm$15.22                      & 68.18$\pm$24.07                 & 65.93$\pm$19.25                           & \textbf{87.18$\pm$13.74}    \\
                                                             &                             &                             & Recall (\%)             & 20.54$\pm$21.64                   & 24.82$\pm$26.31                         & 53.33$\pm$36.70                     & 89.48$\pm$14.37                     & \textbf{94.34$\pm$10.50}          & 75.09$\pm$20.18                      & 73.56$\pm$26.09                 & 88.42$\pm$15.65                           & \underline{92.57$\pm$10.88} \\
                                                             &                             &                             & LE (mm)                 & 57.76$\pm$25.21                   & 41.94$\pm$19.69                         & 3.88$\pm$4.09                       & \underline{1.97$\pm$3.11}           & 2.44$\pm$2.63                     & 5.92$\pm$6.73                        & 4.27$\pm$6.32                   & 4.00$\pm$4.27                             & \textbf{1.27$\pm$2.65}      \\
                                                             &                             &                             & SD (mm)                 & 60.01$\pm$26.43                   & 48.95$\pm$22.78                         & 5.15$\pm$4.22                       & 3.08$\pm$3.83                       & \underline{2.78$\pm$2.67}         & 6.84$\pm$7.25                        & 5.24$\pm$7.15                   & 4.36$\pm$4.21                             & \textbf{1.88$\pm$2.95}      \\
                                                             &                             &                             & nMSE ($\times 10^{-4}$) & 15.14$\pm$5.69                    & 13.18$\pm$7.75                          & 5.21$\pm$3.51                       & \underline{1.94$\pm$1.87}           & 2.04$\pm$3.89                     & 5.19$\pm$7.83                        & 4.41$\pm$3.46                   & 2.96$\pm$2.90                             & \textbf{1.47$\pm$1.54}      \\
            \midrule \multirow{5}{*}{\textbf{15 }}           & \multirow{5}{*}{\textbf{1}} & \multirow{5}{*}{\textbf{2}} & Precision (\%)          & 17.77$\pm$22.52                   & 20.81$\pm$25.53                         & 50.49$\pm$31.46                     & \underline{82.44$\pm$16.59}         & 75.51$\pm$16.60                   & 70.46$\pm$14.81                      & 69.29$\pm$24.25                 & 68.90$\pm$18.63                           & \textbf{87.62$\pm$13.36}    \\
                                                             &                             &                             & Recall (\%)             & 21.09$\pm$25.38                   & 28.10$\pm$29.16                         & 56.64$\pm$36.18                     & 89.65$\pm$14.31                     & \textbf{94.43$\pm$10.30}          & 75.82$\pm$19.84                      & 74.21$\pm$25.99                 & 88.61$\pm$15.26                           & \underline{93.60$\pm$9.97}  \\
                                                             &                             &                             & LE (mm)                 & 37.45$\pm$25.27                   & 26.43$\pm$19.74                         & 4.54$\pm$6.14                       & \underline{1.96$\pm$3.28}           & 2.39$\pm$2.42                     & 5.61$\pm$6.08                        & 4.26$\pm$7.19                   & 3.50$\pm$3.60                             & \textbf{1.14$\pm$1.50}      \\
                                                             &                             &                             & SD (mm)                 & 38.63$\pm$26.50                   & 31.40$\pm$21.26                         & 5.78$\pm$6.16                       & 3.04$\pm$3.89                       & \underline{2.73$\pm$2.45}         & 6.78$\pm$7.13                        & 5.25$\pm$8.20                   & 3.95$\pm$3.68                             & \textbf{1.75$\pm$1.82}      \\
                                                             &                             &                             & nMSE ($\times 10^{-4}$) & 10.79$\pm$6.89                    & 11.52$\pm$6.71                          & 5.59$\pm$5.81                       & \underline{1.94$\pm$1.87}           & 2.04$\pm$4.45                     & 4.97$\pm$7.52                        & 4.13$\pm$3.39                   & 2.70$\pm$2.06                             & \textbf{1.38$\pm$1.47}      \\
            \bottomrule \midrule \multirow{5}{*}{\textbf{5}} & \multirow{5}{*}{\textbf{2}} & \multirow{5}{*}{\textbf{2}} & Precision (\%)          & 11.08$\pm$12.61                   & 13.21$\pm$15.74                         & 51.43$\pm$36.03                     & \underline{81.77$\pm$15.32}         & 79.17$\pm$14.74                   & 62.08$\pm$20.36                      & 59.68$\pm$22.58                 & 64.78$\pm$16.62                           & \textbf{88.30$\pm$12.15}    \\
                                                             &                             &                             & Recall (\%)             & 15.66$\pm$16.57                   & 23.04$\pm$25.38                         & 27.12$\pm$23.14                     & 63.64$\pm$22.20                     & \textbf{70.37$\pm$23.40}          & 51.43$\pm$28.90                      & 55.62$\pm$23.64                 & 64.09$\pm$22.93                           & \underline{67.94$\pm$23.37} \\
                                                             &                             &                             & LE (mm)                 & 81.73$\pm$47.13                   & 69.92$\pm$44.97                         & 3.51$\pm$4.47                       & 2.16$\pm$3.71                       & \underline{2.09$\pm$2.33}         & 6.38$\pm$6.92                        & 8.29$\pm$9.01                   & 4.19$\pm$3.72                             & \textbf{1.23$\pm$2.31}      \\
                                                             &                             &                             & SD (mm)                 & 83.80$\pm$45.57                   & 78.72$\pm$45.84                         & 4.86$\pm$4.59                       & 3.49$\pm$4.15                       & \underline{2.57$\pm$2.26}         & 9.86$\pm$10.67                       & 9.74$\pm$9.86                   & 4.68$\pm$3.95                             & \textbf{1.99$\pm$2.20}      \\
                                                             &                             &                             & nMSE ($\times 10^{-4}$) & 17.11$\pm$7.04                    & 15.22$\pm$7.87                          & 12.59$\pm$75.18                     & 3.89$\pm$3.64                       & \underline{2.89$\pm$2.71}         & 9.47$\pm$7.52                        & 7.57$\pm$4.82                   & 4.93$\pm$3.59                             & \textbf{2.60$\pm$2.65}      \\
            \midrule \multirow{5}{*}{\textbf{5}}             & \multirow{5}{*}{\textbf{3}} & \multirow{5}{*}{\textbf{2}} & Precision (\%)          & 8.52$\pm$9.61                     & 10.75$\pm$12.64                         & 49.88$\pm$35.02                     & \underline{83.54$\pm$15.95}         & 82.83$\pm$13.67                   & 63.87$\pm$17.69                      & 54.76$\pm$24.57                 & 65.58$\pm$17.85                           & \textbf{88.09$\pm$12.01}    \\
                                                             &                             &                             & Recall (\%)             & 13.48$\pm$15.56                   & 20.14$\pm$24.08                         & 22.14$\pm$23.19                     & 60.82$\pm$22.18                     & \textbf{66.81$\pm$23.42}          & 45.91$\pm$31.50                      & 46.61$\pm$22.93                 & 60.90$\pm$22.56                           & \underline{64.82$\pm$22.98} \\
                                                             &                             &                             & LE (mm)                 & 85.81$\pm$48.83                   & 78.75$\pm$47.71                         & 4.44$\pm$5.90                       & 2.27$\pm$4.48                       & \underline{1.67$\pm$1.74}         & 6.17$\pm$6.85                        & 11.19$\pm$10.59                 & 4.80$\pm$4.87                             & \textbf{1.15$\pm$1.70}      \\
                                                             &                             &                             & SD (mm)                 & 88.37$\pm$47.38                   & 89.01$\pm$47.69                         & 5.92$\pm$6.28                       & 3.33$\pm$4.54                       & \underline{2.10$\pm$1.72}         & 8.92$\pm$9.73                        & 13.31$\pm$11.87                 & 5.06$\pm$4.69                             & \textbf{1.75$\pm$1.91}      \\
                                                             &                             &                             & nMSE ($\times 10^{-4}$) & 18.38$\pm$7.57                    & 17.93$\pm$7.92                          & 11.85$\pm$23.20                     & 4.36$\pm$4.20                       & \underline{3.41$\pm$3.98}         & 9.92$\pm$8.70                        & 10.18$\pm$6.27                  & 5.73$\pm$4.43                             & \textbf{3.08$\pm$3.37}      \\
            \bottomrule 
             \midrule \multirow{5}{*}{\textbf{5}}             & \multirow{5}{*}{\textbf{2}} & \multirow{5}{*}{\textbf{3}} & Precision (\%)          & 4.99$\pm$5.14                     & 8.25$\pm$9.43                           & 49.00$\pm$33.36                     & \underline{81.51$\pm$14.54}         & 78.02$\pm$14.09                   & 59.01$\pm$19.24                      & 56.74$\pm$25.98                 & 64.80$\pm$17.21                           & \textbf{85.20$\pm$12.01}    \\
                                                             &                             &                             & Recall (\%)             & 10.17$\pm$15.49                   & 18.17$\pm$22.43                         & 26.55$\pm$21.94                     & 56.73$\pm$21.54                     & \textbf{63.54$\pm$23.55}          & 48.62$\pm$34.16                      & 40.65$\pm$22.43                 & 59.33$\pm$22.53                           & \underline{60.26$\pm$23.03} \\
                                                             &                             &                             & LE (mm)                 & 87.72$\pm$49.98                   & 75.19$\pm$46.91                         & 4.71$\pm$5.84                       & 2.20$\pm$3.53                       & \underline{2.16$\pm$1.89}         & 6.87$\pm$7.22                        & 10.21$\pm$11.09                 & 4.21$\pm$3.55                             & \textbf{1.39$\pm$1.46}      \\
                                                             &                             &                             & SD (mm)                 & 90.20$\pm$48.21                   & 81.78$\pm$47.17                         & 6.28$\pm$6.12                       & 3.59$\pm$3.91                       & \underline{2.68$\pm$1.80}         & 8.06$\pm$9.20                        & 12.48$\pm$12.36                 & 4.74$\pm$3.52                             & \textbf{2.19$\pm$1.72}      \\
                                                             &                             &                             & nMSE ($\times 10^{-4}$) & 19.81$\pm$8.73                    & 17.37$\pm$8.03                          & 11.42$\pm$24.17                     & 5.72$\pm$5.36                       & \underline{4.53$\pm$5.36}         & 10.05$\pm$11.29                      & 11.67$\pm$7.23                  & 6.71$\pm$4.85                             & \textbf{4.32$\pm$4.16}      \\\bottomrule
                                                             \multicolumn{13}{c}{ SimEEG dataset}  \\ 
                                                            \bottomrule \multirow{5}{*}{\textbf{5}} & \multirow{5}{*}{\textbf{{1}}} & \multirow{5}{*}{\textbf{2}} & Precision (\%)          & 12.39$\pm$16.05                   & 12.20$\pm$14.23                         & 43.37$\pm$38.30                     & \underline{78.27$\pm$21.08}         & 70.71$\pm$17.94                   & 62.09$\pm$20.14                      & 54.23$\pm$27.34                 & 59.46$\pm$20.48                           & \textbf{81.94$\pm$18.57}    \\
                                                 &                               &                             & Recall (\%)             & 12.55$\pm$16.12                   & 21.45$\pm$24.63                         & 36.89$\pm$35.82                     & 81.24$\pm$23.28                     & \textbf{92.93$\pm$13.07}          & 65.21$\pm$29.11                      & 70.51$\pm$31.63                 & 88.30$\pm$18.14                           & \underline{89.80$\pm$15.40} \\
                                                 &                               &                             & LE (mm)                 & 50.82$\pm$28.70                   & 43.96$\pm$20.12                         & 3.44$\pm$4.05                       & \underline{2.92$\pm$7.38}           & 3.12$\pm$3.90                     & 6.85$\pm$8.20                        & 6.39$\pm$7.57                   & 5.17$\pm$5.51                             & \textbf{2.03$\pm$4.24}      \\
                                                 &                               &                             & SD (mm)                 & 49.65$\pm$36.75                   & 50.93$\pm$23.59                         & 4.74$\pm$4.20                       & 4.20$\pm$7.70                       & \underline{3.50$\pm$3.87}         & 7.91$\pm$9.18                        & 7.40$\pm$7.98                   & 5.72$\pm$5.69                             & \textbf{2.83$\pm$5.74}      \\
                                                 &                               &                             & nMSE ($\times 10^{-4}$) & 10.53$\pm$3.95                    & 15.17$\pm$7.81                          & 10.36$\pm$41.67                     & 2.68$\pm$2.17                       & \underline{2.35$\pm$2.22}         & 8.99$\pm$8.85                        & 4.71$\pm$3.80                   & 4.15$\pm$3.22                             & \textbf{1.36$\pm$1.66}      \\
            \bottomrule
            
        \end{tabular}
        }
    \vspace{-1.0em}    
    \end{table*}

    \emph{Extendibility on SimEEG dataset} We also evaluate the extendability of \ours~on the SimEEG dataset, which is generated using the same source signal with different head models and lead-field metrics. The results are shown in Table~\ref{tab:simmeg}. As shown in the Table, \ours~still obtains the best precision, LE, SD, and nMSE metrics among the 9 algorithms, which illustrates good extendability for ESI using EEG signals. Note that although a 5\% drop in precision ($87.18\% \rightarrow 81.94\%$) is observed on the SimEEG dataset compared to the SimMEG results, this is understandable as the SimEEG task involves localizing the source from 256-channel signals, whereas the SimMEG dataset utilizes 306-channel MEG signals.

    \emph{Generalizability on Real-world Datasets} Furthermore, we evaluate \ours~ on the two real-world datasets, to see if the proposed method can generalize to more complex tasks in clinical trials. Following~\cite{ref_DeepSIF}, spatial dispersion is used as the metric. As shown in Table~\ref{tab:real-world}, \ours~outperforms other sota algorithms on both of the datasets. This is because that by adaptively refining the feature importance from different views, \ours~learns a better representation from scalp signals, which helps capture key features for ESI.

    \begin{table*}
        [htbp]
        \vspace{-0.5em}
        \centering 
        \caption{Mean Spatial Dispersion of CMR and Localize-MI Dataset. All experiments were repeated 3 times, mean $\pm$ std are reported. The \textbf{best} and \underline{second best} results are highlighted.}
        \label{tab:real-world}
        \resizebox{\linewidth}{!}
        {
        \begin{tabular}{c|cccccccccc}
            \toprule \textbf{Dataset}       & \textbf{sLORETA}~\cite{ref15} & \textbf{Champagne}~\cite{ref_champagne} & \textbf{ConvDip}~\cite{ref_ConvDip} & \textbf{DeepSIF}~\cite{ref_DeepSIF} & \textbf{SSINet}~\cite{ref_SSINet} & \textbf{ADMM-ESI}~\cite{ref_ADMMESI} & \textbf{Catch}~\cite{ref_Catch} & \textbf{FreEFormer}~\cite{ref_FreEformer} & \textbf{\ours}           \\
            \midrule \textbf{CMR}           & 109.68$\pm$32.81                  & 86.31$\pm$27.94                         & 89.34$\pm$51.25                     & \underline{40.38$\pm$21.68}         & 49.81$\pm$30.97                   & 92.58$\pm$31.44                      & 76.32$\pm$26.86                 & 60.17$\pm$25.60                           & \textbf{32.55$\pm$17.19} \\
            \midrule {\textbf{Localize-MI}} & 80.47$\pm$12.35                   & 69.33$\pm$15.71                         & 55.68$\pm$21.84                     & \underline{31.49$\pm$10.82}         & 34.65$\pm$19.30                   & 70.62$\pm$26.57                      & 61.14$\pm$22.09                 & 53.75$\pm$14.35                           & \textbf{22.86$\pm$6.34}  \\
            \bottomrule
        \end{tabular}
        }
        \vspace{-1.5em}
    \end{table*}

    \subsection{Qualitative Results}

\begin{figure}[htbp]
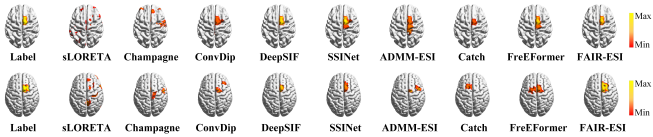

    \vspace{-1.0em}
        \centering
        \begin{minipage}{\linewidth}
            \centering
            \includegraphics[width=\linewidth]{pic/Sim_MEG.jpg}
        \end{minipage}
        \begin{minipage}{\linewidth}
            \centering
            \includegraphics[width=\linewidth]{pic/Sim_EEG.jpg}
        \end{minipage}
        \caption{Visualization of SimMEG and SimEEG datasets, the simulated source is located at region 121. Upper: SimMEG with $[r, N_{s}, N_{n}] = [5, 1, 2]$; Lower: SimEEG with $[r, N_{s}, N_{n}] = [5, 1, 2]$. Colors represent the amplitude of the estimated source activation.}
        \label{fig:combined}
        \vspace{-0.5em}
    \end{figure}

\begin{figure}[htbp]
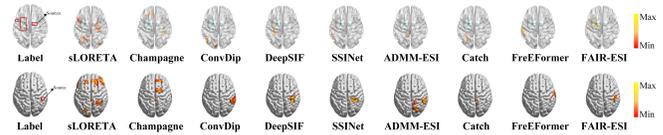

    \vspace{-1.0em}
        \centering
        \begin{minipage}{\linewidth}
            \centering
            \includegraphics[width=\linewidth]{pic/RealMEG.jpg}
        \end{minipage}
        \begin{minipage}{\linewidth}
            \centering
            \includegraphics[width=\linewidth]{pic/RealEEG.jpg}
        \end{minipage}
        \caption{Visualization of CMR and Localize-MI datasets. Upper: CMR dataset; Lower: Localize-MI dataset. Colors represent the amplitude of the estimated source activation.}
        \label{fig:real-world}
    \vspace{-0.5em}
    \end{figure}

    To better illustrate the differences among the 9 algorithms, qualitative results on simulation and real-world datasets are presented in Fig.~\ref{fig:combined} and Fig.~\ref{fig:real-world}.

    \emph{Visualization of Source Localization} As shown in Fig.~\ref{fig:combined}, compared to other methods, \ours~demonstrates the highest consistency with the real labels. Note that while FreEFormer achieves coarse source localization, its inability to precisely estimate signal intensity introduces errors in downstream source signal analysis. Similar results are found on the two real-world datasets. Whereas \ours~can accurately localize the source signal, sLORETA completely fails to localize source signals due to its lack of multi-view refinement for extracted features.

\begin{figure}
            \centering
            \includegraphics[width=0.8\linewidth]{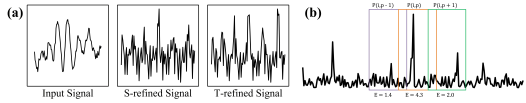}
            \caption{(a) Visualization of spectral-temporal feature refinement; (b)Visualization of key-patch selection, E dnotes the energy of the patch.}
            \label{fig:st-refine}
    \vspace{-1.5em}
    \end{figure}

    \emph{Visualization of Spectral-Temporal Refinement} Fig.~\ref{fig:st-refine} (a) illustrates the features before and after the spectral-temporal refinement. While spectral refinement enhances targeted frequency bands at the expense of signal details, the subsequent temporal refinement mitigates this through weighted feature fusion, ultimately achieving superior feature representation.


    \emph{Visualization of Key-patch selection} As for the patch-refinement, Fig.~\ref{fig:st-refine} (b) presents the energy-based selection of key-patches. From the figure we can find that the spike events generally occupy the highest energy, thus are more likely to be chosen as key-patches. This results in reasonable refinement from a patch-wise perspective.


    \subsection{Abalation Studies}

    \emph{Ablation on Refinement Views} To further evaluate the impact of the feature refinement in each view, we conduct ablation experiments on the SimMEG dataset with $r=5, N_{s}=2, N_{n}=2$. The results are shown in Table~\ref{tab:ablation_module}. As shown in the Table, by combining the three views of feature refinement, \ours~achieves the best precision, LE, and nMSE, with comparable Recall and SD.

    \begin{table}[htbp]
    \vspace{-1.0em}
        \centering
        \caption{Ablation study on feature refinement views. }
        \label{tab:ablation_module} \resizebox{\linewidth}{!}{
        \begin{tabular}{ccc|ccccc}
            \toprule \multicolumn{3}{c|}{\textbf{Refinement Views}} & {\textbf{Precision}} & {\textbf{Recall}}   & {\textbf{LE (mm)}}                            & {\textbf{SD (mm)}}                            & {\textbf{nMSE}}                              \\
            \textit{Spectral}                                       & \textit{Temporal}    & \textit{Patch-wise} & (std)                                         & (std)                                         & (std)                                       & (std)                                       & (std)                                       \\
            \midrule                                                &                      &                     & \makecell{81.22 \\ (13.48)}                   & \makecell{\textbf{68.95} \\ \textbf{(23.46)}} & \makecell{1.79 \\ (1.74)}                   & \makecell{2.32 \\ (1.77)}                   & \makecell{3.02 \\ (2.94)}                   \\
            \midrule \checkmark                                     &                      &                     & \makecell{80.83 \\ 13.94}                     & \makecell{67.63 \\ 23.25}                     & \makecell{1.80 \\ 1.58}                     & \makecell{2.29 \\ 1.63}                     & \makecell{3.04 \\ 2.83}                     \\
            \midrule                                                & \checkmark           &                     & \makecell{80.20 \\ 14.15}                     & \makecell{66.38 \\ 23.20}                     & \makecell{1.84 \\ 1.54}                     & \makecell{2.29 \\ 1.56}                     & \makecell{3.27 \\ 4.16}                     \\
            \midrule \checkmark                                     & \checkmark           &                     & \makecell{\underline{83.99} \\ (13.23)}       & \makecell{67.943 \\ (23.21)}                  & \makecell{\underline{1.53} \\ (1.72)}       & \makecell{2.13 \\ (1.85)}                   & \makecell{2.95 \\ (2.68)}                   \\
            \midrule                                                &                      & \checkmark          & \makecell{82.76 \\ (13.42)}                   & \makecell{\underline{68.05} \\ (23.36)}       & \makecell{1.67 \\ (1.94)}                   & \makecell{\textbf{1.94} \\ \textbf{(2.08)}} & \makecell{\underline{2.97} \\ (2.83)}       \\
            \midrule \checkmark                                     & \checkmark           & \checkmark          & \makecell{\textbf{88.30} \\ \textbf{(12.15)}} & \makecell{67.94 \\ (23.83)}                   & \makecell{\textbf{1.23} \\ \textbf{(2.31)}} & \makecell{\underline{1.99} \\ (2.20)}       & \makecell{\textbf{2.60} \\ \textbf{(2.62)}} \\
            \bottomrule
        \end{tabular}
        }
        \vspace{-1.5em}
    \end{table}

    \section{Conclusion}

    This paper introduces \ours, which improves the precision of ESI by adaptively refining the importance of extracted features from three different views, including \textit{spectral}, \textit{temporal}, and \textit{patch-wise}. Extensive experiments on the simulation and real-world datasets reflect the effectiveness of the proposed method, which provides the potential for further evaluations on the functionalities of the brain and the mechanism of brain disorders.

    \bibliographystyle{unsrt}
    \bibliography{ref}
\end{document}